# Dimensionality Reduction and Motion Clustering during Activities of Daily Living: 3, 4, and 7 Degree-of-Freedom Arm Movements

Yuri Gloumakov, S*tudent Member*, *IEEE*, Adam J. Spiers, *Member*, *IEEE*, and Aaron M. Dollar, *Senior Member*, IEEE

*Abstract*—The wide variety of motions performed by the human arm during daily tasks makes it desirable to find representative subsets to reduce the dimensionality of these movements for a variety of applications, including the design and control of robotic and prosthetic devices. This paper presents a novel method and the results of an extensive human subjects study to obtain representative arm joint angle trajectories that span naturalistic motions during Activities of Daily Living (ADLs). In particular, we seek to identify sets of useful motion trajectories of the upper limb that are functions of a single variable, allowing, for instance, an entire prosthetic or robotic arm to be controlled with a single input from a user, along with a means to select between motions for different tasks. Data driven approaches are used to obtain clusters as well as representative motion averages for the full-arm 7 degree of freedom (DOF), elbow-wrist 4 DOF, and wrist-only 3 DOF motions. The proposed method makes use of well-known techniques such as dynamic time warping (DTW) to obtain a divergence measure between motion segments, DTW barycenter averaging (DBA) to obtain averages, Ward's distance criterion to build hierarchical trees, batch-DTW to simultaneously align multiple motion data, and functional principal component analysis (fPCA) to evaluate cluster variability. The clusters that emerge associate various recorded motions into primarily hand start and end location for the full-arm system, motion direction for the wrist-only system, and an intermediate between the two qualities for the elbow-wrist system. The proposed clustering methodology is justified by comparing results against alternative approaches.

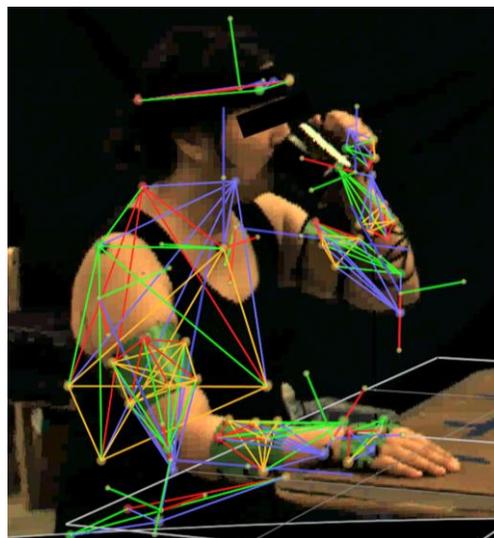

Fig. 1.  Subject performing an ADL task, drinking from a mug. The subject's motion capture 'skeleton' is superimposed in this image. Redundant markers are included to enable the prediction of occluded marker locations and maintain the ability to identify joint centers.

## I. Introduction

THE human arm is a remarkable tool that affords us the ability to accomplish complex manipulation tasks. Unlike the study of the lower limbs with regard to gait, the arm has much more varied patterns of motions that it regularly performs [1]. Despite this, humans consistently perform various reaching, grasping, and manipulation tasks in a relatively predictable pattern [2] without much cognitive burden. Since there exists some apparent regularity of human motion patterns despite the huge space of possible motions, it is predicted that simplified models of the motion can be found, for example, by extracting a subset of representative movements. We investigate a data driven clustering approach to identify natural groupings of the 7 degree-of-freedom (DOF), 4 DOF, and 3 DOF joint angle trajectories of the upper-limb, elbow-wrist, and wrist-only, respectively (hereafter simply referred to as "arm motions"), obtained from individuals performing a range of selected activities of daily living (ADLs). We ultimately seek to find a relatively small set of "useful" arm motion trajectories that are a function of a single variable. This approach would, for instance, allow an upper-limb amputee to control a multi-DOF prosthetic arm using a single control input, such as from two-site EMG, which is the current standard in clinical practice [3].

Reduced dimensionality representations of upper-limb movements are useful in a variety of domains, including controlling the functionality of a semi-autonomous prosthetic

This paragraph of the first footnote will contain the date on which you submitted your paper for review. This work was supported by the Congressionally-Directed Medical Research Programs (CDMRP) under grant-W81XWH-15-10407. The work presented in this paper is an expansion of the authors' previous work found in [19].

Y. Gloumakov and A.M. Dollar are with the Mechanical Engineering and Materials Science Department, Yale University, New Haven, CT 06511 USA, (email: {yuri.gloumakov, aaron.dollar} @yale.edu).

A.J. Spiers is with the Max Planck Institute for Intelligent Systems, Stuttgart, 70597, Germany (email: a.spiers@is.mpg.de).

device by using sub-motions to recreate a larger set of possible tasks. Research groups investigating joint synergies to control active prosthetic wrists or elbows have primarily focused on reaching motions [4], [5]. While our methodology is not limited to only this application, the development of an arm motion hierarchy formalizes the stratification of reaching and manipulation; enabling researchers to evaluate various degrees of motion specificity.

Out of the infinitum of motions that the human arm can achieve, we looked to only use the most useful ones across individuals, i.e. most common ADLs, as the set of motions to cluster (Fig. 1). For the tasks we asked our subjects to perform in this work, we selected ones largely inspired by the standardized 'outcome measure' arm function assessment tools of AM-ULA [6] and various surveys that queried motion-impaired participants on common tasks that they find difficult [7]–[9]. These tasks generally relate to food preparation, eating, hygiene, grooming, and dressing, and are crucial for independent living.

Past research on upper limb motion has spanned a variety of fields with different research groups exploring various techniques to extract insight into how humans control and make use of their upper limbs. Such research has covered non-linear control, neural networks, and musculoskeletal modelling [10]. Some groups have also attempted to identify and make use of underlying healthy motion patterns to control upper-limb prosthetic devices. These investigations include using artificial neural networks to predict or discriminate upper-limb functions [5], [11] or performing pattern recognition of simultaneous motion primitives [12] in healthy subjects. Other groups examined healthy participants performing various tasks and extracted a subset of arm motion primitives using functional principal component analysis (fPCA) [13]–[15]. Instead of using a linear combination of movement primitives to construct a motion, a much more straight forward approach to controlling an upper-limb device could instead focus on clusters of sequential sub-motions that recreate the complete task, as is proposed in this paper. On-line motion recognition, as well as a hierarchical description of non-ADL motion segments has been performed in [16]. However, the focus was on automatic motion recognition of the whole body rather than on sequential motion segments and results were not deterministic. Other related fields include rehabilitation efforts, which have investigated motion patterns of healthy participants by analyzing only the ranges of joint angles [17], [18]. Therefore, although some groups have attempted to extract underlying simplified motion patterns [5], [12], [13], [16], none have used a clustering approach that stratifies arm motions related to ADLs.

This paper is an extension of a previous conference paper by the authors [19], and expands and extends it in a number of ways. It examines 4- and 3-DOF cases in addition to 7-DOF, increases the number of subjects (from 5 to 12), establishes a set of motion modalities for each DOF model, analyzes the variabilities in motion within each of the clusters using fPCA, and demonstrates results using accompanying animations visually reassuring their use in real-world applications. One of the findings in [19] was that 7 DOF clusters primarily relied on task location, in other words end effector location seemingly dominated the results over hand orientation. Therefore, we further investigate this by directly clustering the 4 DOF shoulder and elbow joint angle trajectories without the wrist.

The additional DOF models are primarily analyzed for application in technologies assisting patients with different degrees of arm disability or amputation (i.e. full arm (7 DOF), elbow and wrist (4 DOF), and wrist only (3 DOF)). This includes transradial amputations (artificial wrist and terminal device only), transhumeral amputations (artificial elbow, wrist, and terminal device), and shoulder disarticulation and higher (artificial shoulder, elbow, wrist, and terminal device).

TABLE I
TASKS AND CORRESPONDING MOTION SEGMENTS

| Task Code* | Standing Tasks** |
|---|---|
| t2b | (1) reach for box on top shelf (2) move box to bottom shelf (3) return hands |
| b2t | (1) reach for box on bottom shelf (2) move box to top shelf (3) return hands |
| t2m | (1) reach for box on top shelf (2) move box to middle shelf (3) return hands |
| m2t | (1) reach for box on middle shelf (2) move box to top shelf (3) return hands |
| m2b | (1) reach for box on middle shelf (2) move box to bottom shelf (3) return hands |
| b2m | (1) reach for box on bottom shelf (2) move box to middle shelf (3) return hands |
| ke | (1) bring key to keyhole (2) turn key (3) turn back (4) remove key from keyhole and return hand |
| kn | (1) reach for door knob (2) turn knob (3) turn back (4) return hand |
| dh | (1) reach for door handle (2) open door (3) return hand |
| oh | (1) reach for can on top shelf (2) bring can down in front of the body |
| mp | (1) reach for mug in location C1 (2) take a sip (3) return mug (4) return hand |
| md | (1) reach for mug in location C2 (2) take a sip (3) return mug (4) return hand |
| mc | (1) reach for mug in location C3 (2) take a sip (3) return mug (4) return hand |
| cp | (1) reach for cup in location C1 (2) take a sip (3) return mug (4) return hand |
| cd | (1) reach for cup in location C2 (2) take a sip (3) return mug (4) return hand |
| cc | (1) reach for cup in location C3 (2) take a sip (3) return mug (4) return hand |
| st | (1) reach for suitcase (2) transfer suitcase to table (3) return hands |
| ax | (1) bring hand to contralateral axilla (2) return hand |
| pt | (1) bring hand to back pocket (2) return hand |
| | Sitting tasks** |
| sp | (1) reach for spoon (2) bring spoon to bowl (3) scoop (4) bring to mouth (5) return spoon (6) return hand |
| fr | (1) reach for fork (2) stab the middle of the plate (3) bring to mouth (4) return fork (5) return hand |
| ms | (1) reach for mug (2) take a sip (3) return mug (4) return hand |
| cs | (1) reach for cup (2) take a sip (3) return cup (4) return hand |
| pr | (1) reach for cup (2) pour into another cup (3) return cup (4) return hand |

*Task codes are used in the results section
**Unless otherwise specified, standing tasks started and ended with the subjects' hands by their side while for sitting tasks the hands were to start and end on the table palm side down.

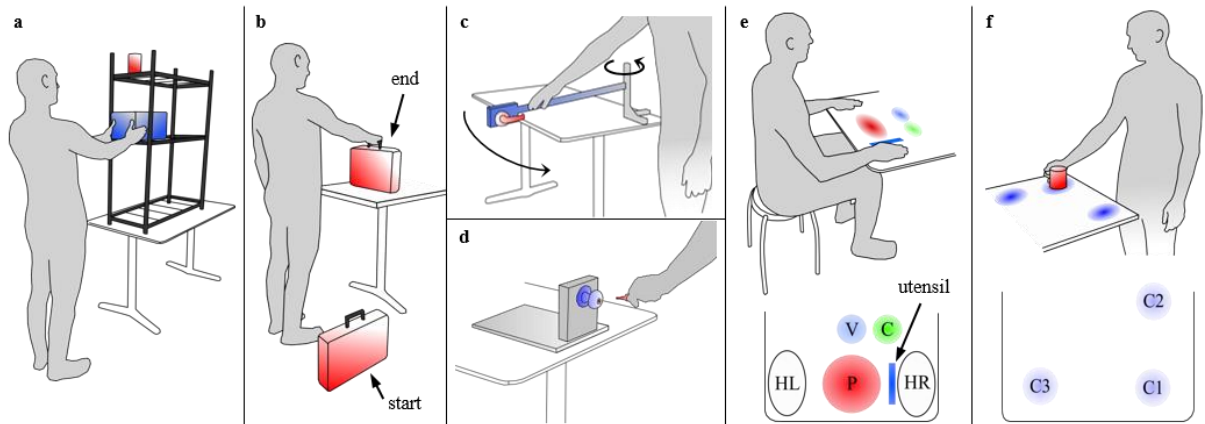

Fig. 2. Depictions of several selected protocol tasks: (a) a box object was to be moved from one specified shelf to another. The object on the top shelf is the location of the can during overhead reaching tasks. (b) The initial and final locations of the suitcase tasks, (c) simulated door opening task, and (d) simulated door knob and key tasks. (e) The set up for the sitting tasks: the left and right hand start and end in HL and HR, a utensil is placed next to HR, a bowl or plate are placed in P, a cup or mug is placed in C, and a container to collect the water during the pouring task is placed in V. (f) The three target locations of the standing cup and mug tasks, during which the table is elevated to simulate a countertop, where C2 is 25 cm from C1 and C3 is 45 cm from C1. The task conditions for left handed participants are mirrored. Table height is 74 cm, and is elevated to 92 cm to simulate a counter top for the standing cup and mug tasks. The mug (9.5 cm height, 8 cm diameter), can (7.5 cm height, cm diameter), box (21x37x19 cm), and suitcase (43x9x30 cm) weigh 0.36, 0.09, 0.23, and 1.36 kg respectively. The shelves are 80, 140, and 180 cm above the floor. Door knob and handle are 90 cm above the floor, and the simulated door swivels with an 84 cm radius.

## II. EXPERIMENTAL PROTOCOL

### A. Task Protocol

The set of motions that are used in this study were collected from healthy individuals performing tasks that generally occur during daily life. The tasks used in the study, which were based on the standard functional measure AM-ULA [6], are listed in Table 1 with the setup described in more detail in Fig. 2. We only included a subset of tasks found in AM-ULA that naturally could be segmented into sub-motions, which is important for analyzing distinct motion segments related to ADLs rather than an entire complex motion that occurs during a task. For example, the task of drinking from a cup may involve clear segments of reaching, grasping, bringing to the mouth, and returning to a table. Tasks such as folding a towel or putting on a shirt were omitted from the protocol due to lack of distinct motion segments. Small amplitude cyclical tasks such as cutting with a knife or stirring were also omitted.

The protocol was completed by 12 healthy subjects (6 male, 6 female) who performed the 24 tasks 3 times each, to provide a way to average or smooth the motions during analysis as well as to account for outliers. For results to be as generalizable as possible, participants were additionally chosen to span 24 to 71 years of age. Each task was segmented into 2 to 6 distinct sub-motions, totaling to 85 motion segments per person. Each participant performed the protocol over the course of 5 hours in a single visit. They were instructed to start and end each task in specified 'rest poses', i.e. standing with hands by their sides or sitting with palms on a table surface. Minimal additional instruction were given on how to perform the task. Experimental set-up was inverted for left-handed participants.

This study protocol was approved by Yale University Institutional Review Board, HSC# 1610018511.

### B. Data Acquisition

Motions were recorded with a Vicon Motion Capture System (Oxford Metrics Limited, Oxford) using 12 infrared 'Bonita' model cameras, 1 video reference camera (synchronized with the Vicon system), and 55 body-worn reflective markers at a rate of 100 frames/second. Synchronized video from the reference camera was used to aid in marker identification in the Vicon Nexus software.

## III. DATA ANALYSIS

The goal is to identify how upper-limb motions related to ADL cluster and obtain a subset of representative motions using data driven approaches. The data processing and analysis pipeline is illustrated in Fig. 3. Collected data is first processed in the segmentation step in which each task (which was recorded as a separate motion capture file) is manually split into sequential reaching and manipulation joint angle trajectories. Each task sub-movement is averaged across individuals and repetitions to curb the influence of outliers during the clustering phase. A divergence measure is chosen such that it reliably computes a similarity measure between motion segments, which are followed by a clustering step. The clusters are evaluated twice: first to decide on the number of clusters, and second against alternative algorithms using an original scoring metric to validate the chosen methodology. Finally, representative motions are obtained from each cluster by averaging and their respective variances are computed. Since the 4 DOF shoulder-elbow system is included solely to compare against the 7 DOF system, this portion of the analysis is limited to only obtaining the clusters.

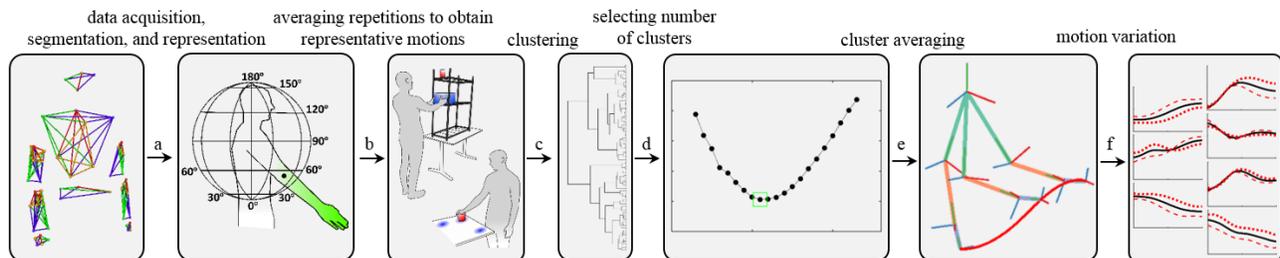

Fig. 3. General framework of the data processing and analysis. (a) Cartesian coordinates of markers tracking human motion are converted to arm joint angles, creating a set of feature variables generalizable across subjects. (b) Repetitions of different motions and subjects are segmented and averaged. (c) The motions are compared using DTW and clustered using agglomerative hierarchical clustering with Ward's linkage distance. (d) The L method is used to select the number of clusters from the dendrogram. (e) Each cluster is averaged and (f) within cluster variations are calculated using fPCA. Steps (b-f) are repeated for each of the three DOF arm models. Steps (b-d) are repeated once more for the 4 DOF shoulder-elbow model.

### A. Motion Representation

Human arm motion data can been described in various ways, such as using Cartesian coordinates of the humerus, forearm, and hand, or joint angles obtained from the shoulder, elbow, and wrist. For motion reconstruction or down-sampling, the joint angle method suffers from the unequal impact that different DOF have on the end effector trajectory. However, fewer variables are required to reconstruct the upper-limb using joint angle definitions. This is an important factor when calculating the similarity between motions, and is easily interpretable and implementable in prosthetic devices. The simplicity of the joint-angle system is therefore used through the rest the paper.

The upper-limb joint angle systems analyzed are based on 7 DOF shoulder-elbow-wrist, 4 DOF elbow-wrist, and 3 DOF wrist-only definitions according to [20], hereby referred to simply as 7 DOF, 4 DOF, and 3 DOF models, respectively. Additional analysis is performed on the 4 DOF shoulder-elbow as well. The shoulder angles consist of plane of elevation, angle of elevation [21], and internal axial rotation, using the second option for the humerus coordinate system in [20] and is detailed in Fig 4. The elbow angle is considered between the forearm and humerus, while wrist angles consist of supination, wrist flexion, and hand deviation. For left-handed participants, the joint angles were inverted so that they are congruous to right-handed participants.

### B. Motion Segmentation

Arm motions during ADL can be seen as a composite of individual sub-motions with which generalized tasks, such as drinking from a cup, are accomplished. Quantitative approaches to segmentation include derivative or zero velocity threshold [22], principle component analysis (PCA) [23], or a hybrid Hidden Markov Model (HMM) and PCA approach [24]. Despite advances in the field, verification of the segmentation algorithms were generally performed by comparing to a heuristically defined ground truth. Therefore, for the purposes of the present work, we segmented the motions manually each time the end effector reached zero velocity; when the participant made contact with, acquired the food item (analogous to [25]), transferred, returned the object, completed the task, or returned the hand back to its 'rest pose', detailed in

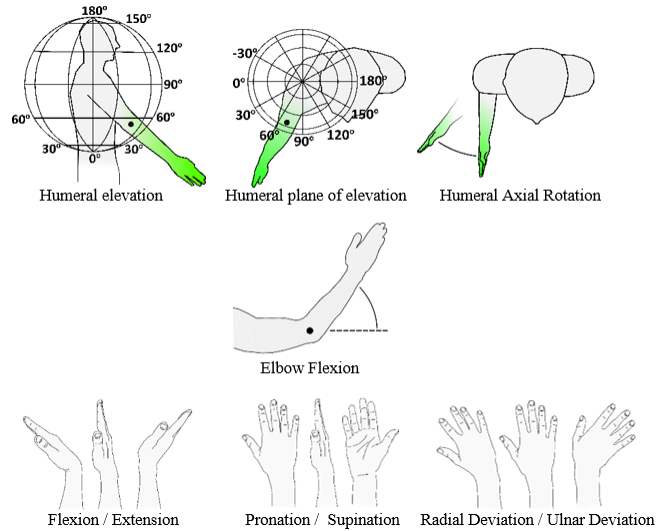

Fig. 4. Humeral elevation and plane of elevation are depicted using the globe system described in [15]. The elbow is positioned below the shoulder in the image to depict humeral axial rotation.

Table 1.

### C. Divergence Measure

Obtaining a divergence between time-series data requires that the data, or the corresponding feature vectors, be of equal length. While resampling or modeling time-series data frequently leads to a loss of some information, dynamic time warping (DTW) does not [26]. DTW works by replicating the frames between two time-series such that it minimizes the sum of square Euclidean distances while simultaneously making them equal in length. Divergence is calculated by summing the distances between each pairs of points of the two newly warped trajectories. It works according to the following equation,

$$D(i,j) = min \begin{Bmatrix} D(i-1,j) + d(i,j) \\ D(i-1,j-1) + d(i,j) \\ D(i,j-1) + d(i,j) \end{Bmatrix}, D(1,1) = d(1,1)$$

(1)

where $d(i,j)$ corresponds to the Euclidean distance between the DOF of frame $i$ of one motion segment and the DOF of frame $j$ of the second motion segment. The optimal path is then calculated through matrix $D(i,j)$ by starting at the last frames of each of the motions and moving backwards through the smallest distance values.

In order to capture the distance between arm motions that might be moving in opposite directions, such as bringing a cup to the mouth and returning the cup back to the table, DTW between each pair of motions was calculated twice: once with the original motion data, and once with one of the motions going in reverse. The smaller of the two divergence values is saved for the clustering step. Divergence values are normalized by dividing by the new time duration obtained during DTW. This is done so that the DTW comparison made between longer motion segments is comparable to shorter motion segments, and we refer to it as normalized-DTW.

### D. Averaging Motions

Averaging of motions is performed during two separate phases throughout the analysis. The first time it is to average repetitions from multiple participants; an average for the same sub-motions from the same task across all individuals was computed prior to clustering. Each average included 36 motion segments, three from each participant. Averaging also had a beneficial effect of minimizing the impact of noise and outliers; something hierarchical clustering is particularly sensitive to. The second use of averaging is to identify a representative motion for each cluster. A time-series average can be obtained in a variety of ways, one of which is linearly resampling all the data to the same length and taking a frame by frame average. This approach is sensitive to phase shifts, where motion epochs are poorly well-aligned, so instead we used a DTW barycenter averaging (DBA) algorithm [27].

One precaution that had to be made during DBA is that it is prone to local minimums, where the consensus segment will accentuate the amplitude of certain frames to minimize the DTW distance [27]. Although more complex algorithms exist that attempt to deal with such issues, such as [28], we simply limited the amount of frames that can be warped to the minimum amount possible when performing DTW between the shortest and the longest motion segment pair in each group.

### E. Agglomerative Hierarchical Clustering

An effective clustering algorithm minimizes variation within clusters while maximizing the differences between clusters, while also describing the overall structure of motions. Unlike CURE [29] or Chameleon [30], the algorithm would preferably create relatively "spherical" clusters, in which the arm motions in a cluster are more similar to one another than to motions belonging to other clusters. Agglomerative hierarchical clustering [31] with Ward's linkage criterion, or simply distance, accomplishes this while providing an easily interpretable dendrogram depicting how clusters are formed and their relationship to one another. It works by successively merging individual motions based on the shortest specified pairwise divergence into a single cluster until one cluster containing all of the data is left. Ward's linkage criterion is preferable over other linkage algorithms, such as complete linkage (or furthest-neighbor) or single linkage (or nearest-neighbor), for this application as it creates distinct and relatively spherical clusters by accounting for both the within and cumulative cluster variances according to

$$W = SS_{12} - (SS_1 + SS_2) \qquad (2)$$

where $SS_1$ and $SS_2$ are the sum of squares of each of the members of the cluster to its respective centroid, and $SS_{12}$ is the sum of squares of the combined cluster. $W$ is the calculated Ward's distance value. This computation is performed for each subsequent cluster without the need to identify the center of the clusters directly. One of the downsides to using this algorithm is its inability to adjust once a merge decision has been executed [32]. Thus, as described in Section D, we hope that outlier effects are largely mitigated prior to clustering.

A set number of clusters can be extracted in a variety of ways from the dedrograms. While heuristics can be used to select a seemingly reasonable number of clusters for the 7 DOF model using intuition, the 4 DOF and 3 DOF models do not lend themselves to an easy interpretation. Therefore we use a data driven approach called the L method [33] to identify an "optimal" number of clusters. The method is used with a greedy evaluation approach, as recommended by [33] and only considers the Ward's distance (2) value between the two clusters being merged. Unlike other approaches that only evaluate the data locally or are sensitive to noise, the L method makes use of the entire set of distance values between each merging pair to determine the point of transition between the internally homogenous and non-homogenous cluster merging phases. It works by linearly fitting each phase while varying the sequence of points that belong to each and calculating the total error, $RMSE_{tot}$, according to

$$RMSE_{tot} = \frac{c-1}{b-1} \times RMSE(L_c) + \frac{b-c}{b-1} \times RMSE(R_c) \qquad (3)$$

where $c$ and $b$ correspond to the partitions of the distance data belonging to the left and right side, respectively, and $L_c$ and $R_c$ are the lines of best-fit, respectively. $L_c$ and $R_c$ must have at least two points, and $c$ and $b$ always add up to the total number of points. A value of $c$ which minimizes $RMSE_{tot}$ corresponds to the "optimal" number of clusters, and is called the "knee". Certain improvements to the L method were additionally recommended by the authors [33], and are implemented in the results. These include adjusting the number of mergings that are being evaluated and removing the set of data left of the point corresponding to the largest merging distance.

### F. Cluster Quality

By re-computing the hierarchical clustering dendrogram using individual motions, rather than the average of each motion type, we can compute an evaluation score that captures how consistently repetitions cluster. For every pair of the same motion segment from the same individual that is clustered together a score is increased by one point. The quality of clustering is then calculated by taking the score for a set number of clusters and dividing it by the maximum possible score, only obtained when repetitions belonging to the same subjects are clustered correctly. It follows that a single cluster of all data receives a perfect quality score that monotonically decreases with an increased number of clusters. The evaluation score could theoretically remain at 100% up to 1020 clusters; 85 unique motions from 12 participants. Common clustering methods are additionally evaluated to validate the selection of the primary methodology: K-medoids clustering [34] and Euclidean distance between motions represented using coefficients belonging to cubic Bézier fits.

K-medoids clustering is tested using DTW divergences between motions. Unlike K-means, K-medoids identifies a median motion segment instead of calculating a centroid. At each iteration distances between the representative cluster object and all other motions are calculated using DTW, cluster membership is updated, and a new cluster median is found. This algorithm was performed ten times to curb local minimums.

To test an alternative divergence measure, cubic Bézier curves are fit to each joint angle trajectory using least squares, yielding a set of Bézier control points that represent each motion segment. Cubic Béziers have been shown to accurately represent human motion during data compression [35] and hand trajectories [36], and are therefore chosen. One benefit to using Bézier curves over traditional polynomials is that the first and last control points correspond to the start and end locations of a trajectory. Since cubic Bézier curves are used, the feature vectors are therefore 28, 16, and 12 elements long for the 7 DOF, 4 DOF, and 3 DOF models, respectively, corresponding to 4 control points. A divergence measure is obtained by calculating the Euclidean distance between motion segments using the feature vectors. The segments are then clustered using hierarchical clustering with Ward's linkage criterion.

*G. Within Cluster Average and Variation*

In order to obtain a sense of motion variation within each cluster, an average motion was first found, motions were resampled to be equal in duration, and fPCA [37] was used to extract the principle components. Each set of the first $n$ principal components then explains some amount of variation in motion data. Clusters with a lot of motion variability will require more principal components to explain the same amount of variation than clusters with relatively homogenous segments.

As described in section II. A., each motion within a cluster is an average of 36 individual motion segments, therefore a cluster with 2 motions can also be analyzed as a set of 72 individual motion segments. All of the individual motions that occur while replacing the object or returning the hand are first reversed. Then, as in section III. D., DBA is used to identify the average of each cluster, initializing it to have the same number of frames as the longest motion. The individual motions are then resampled to equal in length using batch-DTW [38]. Unlike linear resampling, batch-DTW is better suited in this application by aligning epochs independently for each motion, thus better capturing motion variability. Batch-DTW is an asymmetric DTW algorithm which simultaneously aligns multiple time-series data and retains a non-increasing time-duration, something that is impossible to achieve using standard DTW. It works by first selecting a reference time-series segment, in our case it is the average motion of a cluster, and performing DTW with each of the other time-series data. Each set of frames that are repeated for the reference segment, the other segment has those frames averaged instead. An example would be if the optimal warping path included *(i-1,j)*, *(i,j)*, *(i+1,j)*, where the *(i-1)th*, *ith*, and *(i+1)th* frames of motion $M_i$ is aligned with the *jth* frame of the reference motion $M_j$. Batch-DTW would take the following average of the three frames

$$\left(M_i(i-1,:) + M_i(i,:) + M_i(i+1,:)\right) / 3$$

Three 3rd order B-Spline [39] elements were fit to each of the newly aligned motion segments (using least squares). The coefficients of the curves are used as feature variables when calculating the principle components [37]. The coefficients obtained from the principle components can then be used to reconstruct the curves of variability around the average motion. Since the motion alignment considers only the positions of the joint angles, velocity and acceleration information is lost, therefore instead of a 5th order fit as recommended in [40], 3rd order was chosen instead. Three equally spaced B-spline elements were primarily used to better capture the start, middle, and end phases of the joint angle trajectories.

IV. RESULTS

Fig. 5 displays dendrograms obtained for the joint angle 7 DOF full-arm model, 4 DOF elbow-wrist model, 3 DOF wrist-only model, and the 4 DOF shoulder-elbow model. A horizontal cut is used to segment each of the dendrograms to obtain a subset of clusters according to the L method described in [33] using the greedy approach, whose results accompany the dendrograms in Fig. 6. The L method identified the following set of clusters: 5 clusters for the 3 DOF model, and 11 clusters for the rest. The shoulder-elbow trajectory dendrogram is nearly identical to the 7-DOF model barring two motions being placed in difference clusters, st-2 (transfer suitcase to table) and fr-2 (use fork).

One of the L method adjustments recommended by the authors [33] was to dynamically adjust the number of mergings being evaluated down to a minimum of 20 points. In our case, the identified "knee" for 25 merging points was equivalent and we therefore left the additional 5 points in. The largest merging distance for each DOF model was the first merging and therefore the data being evaluated started with the merging distance between 2 and 3 clusters.

Evaluation of the chosen methodology is shown against an alternative divergence measure and clustering algorithm while varying the number of clusters from 1 to 25 (Fig. 7). This was done for each DOF model. The chosen clustering methodology consistently outperforms the other methods for almost every number of clusters.

Due to practical limitations in representing multi-DOF motion with images or complex equations, we include all of the resulting average motions and the first two principle components of each cluster in the multi-media accompanying this paper. An example average motion representing the 8th cluster of the 7 DOF model, *reach-to-front-far,* is shown in Fig. 8, in which the start, middle, and end poses of the arm are displayed. The location of the end effector is also traced out throughout the motion. The stick model is created using forward kinematics of the average motion's DOF in MATLAB (MathWorks, US) according to [20], and the accompanying skeleton model was created using an online skeletal animation tool, KineMan (http://www.kineman.com). The first principle component for each DOF of the motion is also included in the figure. Start and end locations of the average of the 4th cluster from the wrist model, *supination + flexion*, are additionally shown in Fig. 9. The motions for the wrist and elbow-wrist models were depicted using only the KineMan tool.

Variation of the motions within each cluster is captured using fPCA. The percent of the variability explained by each set of principal components, i.e. the first n number of principal components, is summarized in Fig. 10. For each cluster the average pair-wise divergence between cluster members is additionally included, calculated using normalized-DTW. The analysis indicated that while some clusters needed only 3 principal components to describe 80% of the variation, others needed as many as 8.

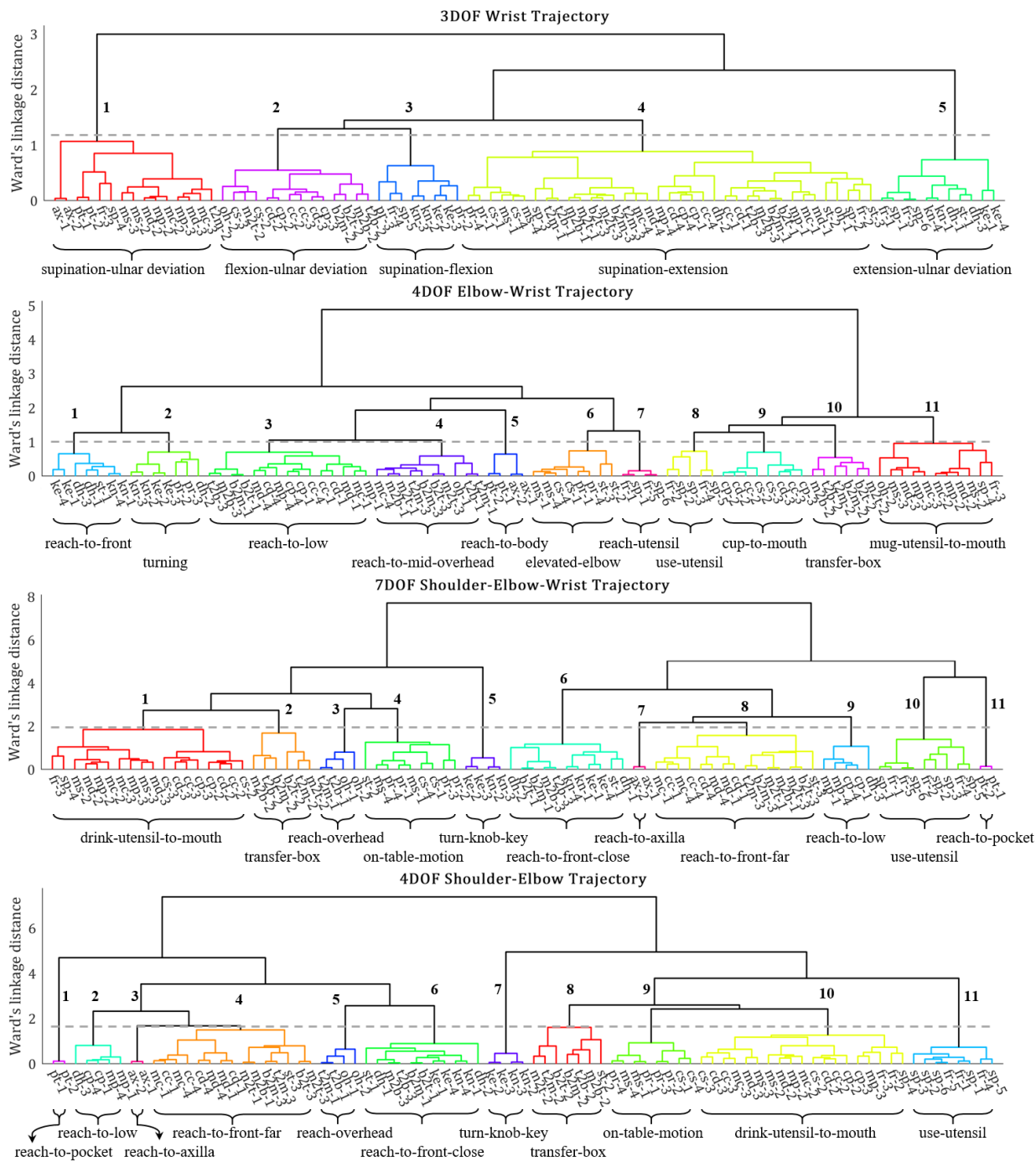

Fig. 5. Dendrograms for the 3, 4, and 7 DOF models. Location of the horizontal cut (dashed line) was chosen using results of the L method. An appropriate cluster name accompanies each of the clusters: major axes of wrist rotation for the 3 DOF model and generalized description of the motions for the 4 and 7 DOF models. Cluster colors are auto-generated and are unrelated between dendrograms.

## V. DISCUSSION

Although the hierarchical tree does not output a specific number of clusters, clustered groups can be obtained by transecting the dendrogram at a desired value. The most straightforward method is using a straight line cut as is seen in Fig. 5. The location of this cut was chosen using a data driven approach called the L method with greedy evaluation, chosen over global primarily due to greater reliability when selecting the number of clusters [33]. Global evaluations have shown only minor deviations and were not considered in the analysis.

According to the L method, unlike for the 4 DOF elbow-wrist model, 7 DOF and 3 DOF models have a clear RMSE minimum suggesting 11 and 5 clusters, respectively. Clusters obtained for

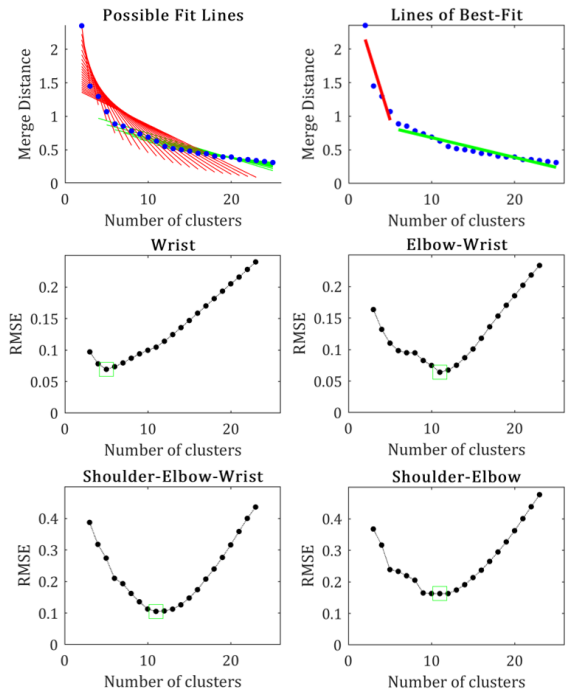

Fig. 6. L method results for each of the models. An example of the identified "knee" for the Wrist model is included at the top row.

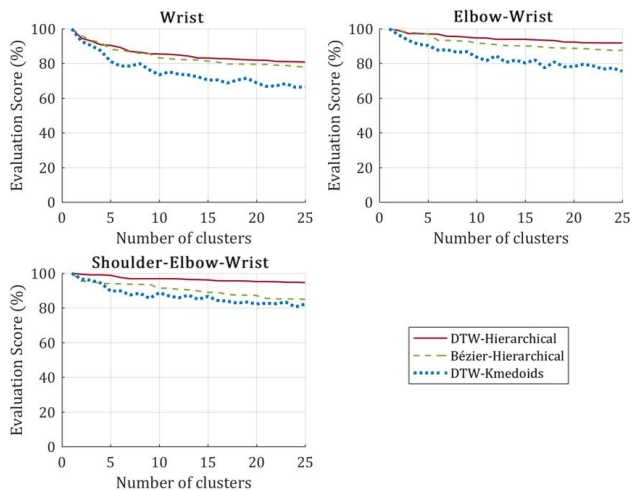

Fig. 7. Quality of clustering for different divergence measures and clustering algorithms across a range of number of clusters. Scoring metric assessed how frequently repetitions from the same individuals clustered together.

the 7 DOF model, similar to results found in our previous work [19] and consistent with the spatial control hypothesis [41], are predicated on hand start and end locations while smaller groupings within each cluster are based on other movement characteristics. This suggests that either the wrist motion is synergistic with the shoulder and elbow joints along the motion path [5], [42], or that its range of motion was not significant enough to influence clustering. Depending on the set of motions being studied, it is likely that both are factors. To test this we analyzed the shoulder-elbow trajectories, which identified nearly identical clusters to the 7 DOF model, confirming that arm motions primarily clustered according to task location. This suggests that when designing a 7-DOF prosthetic device control scheme, priority should be given to the location of the end effector. The 3 DOF model also created clusters primarily based

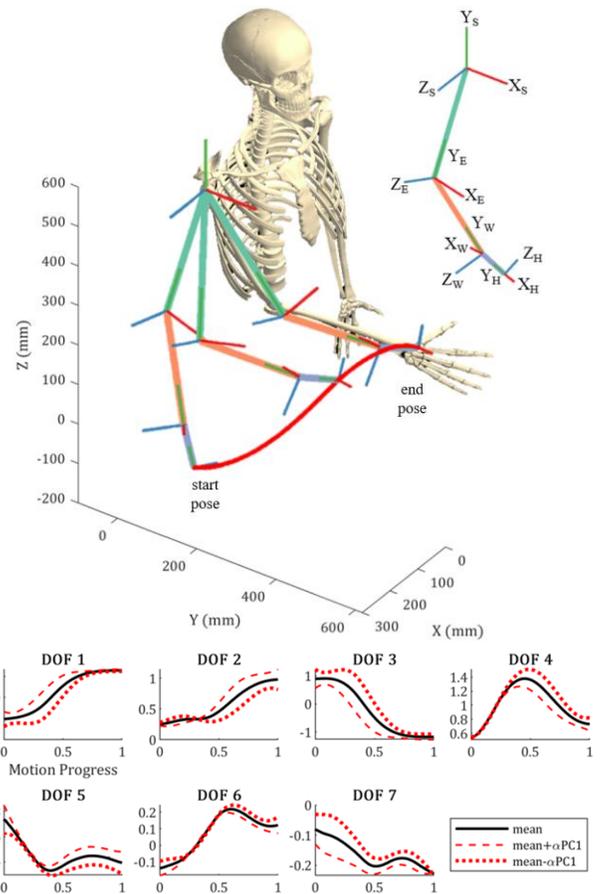

Fig. 8. Forward kinematics are used to display the average motion of the 8$^{th}$ cluster for the 7 DOF model, *reach-to-front-far*. Three reference frames are displayed with X, Y, and Z axis using subscripts S, E, W, and H for shoulder, elbow, wrist, and hand, respectively. The shoulder coordinate frame is fixed throughout the motion. Humerus, forearm, and hand lengths correspond to an average adult. DOF angle correspond, respectively, to humeral elevation, plane of elevation, internal rotation, elbow flexion, wrist supination, wrist, flexion, and wrist deviation. Individual joint angle trajectories are displayed along with the first principal component. $\alpha$ was set to equal the proportion of total variation explained by that component.

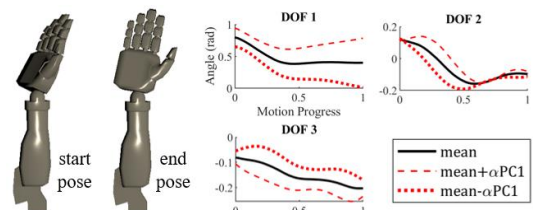

Fig. 9. Start and end poses of the 4$^{th}$ cluster for the 3 DOF model, *supination+ flexion*, are shown on the left along with the joint angle trajectories and the first principle component on the right. The three joint angles in order correspond to supination, flexion, and deviation. $\alpha$ was set to equal the proportion of total variation explained by the principle component.

on starts and ends of the wrist joint angle trajectories.

Although the global minimum is located at 11 clusters, the 4 DOF elbow-wrist model has an additional RMSE minimum at 6 clusters, indicating the possibility of a second plausible interpretation: clustering result for the 4 DOF model is not a gradual transition between the 7 DOF and 3 DOF models, but rather it exhibits both of their minimums simultaneously. We therefore suspect that 11 and 6 cluster minimums correspond to hand location and wrist orientation, respectively. Although the dendrogram structure for the 4 DOF model is more difficult to interpret, given that 11 clusters were ultimately identified despite the absence of shoulder angles, it would appear that task location information is largely maintained in the elbow trajectory, consistent with the efforts in [5].

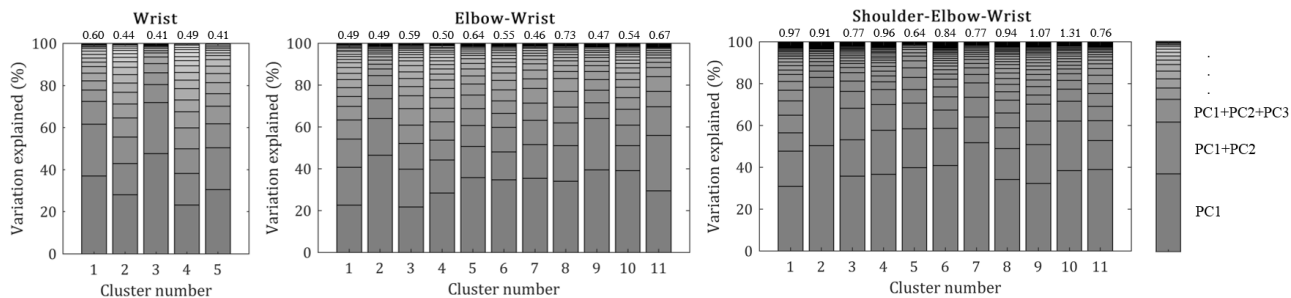

Fig. 10. The variation explained by each set of principal components for each joint angle system's averages are displayed. Note that clusters requiring more principal components to explain the same amount of variation is generally consistent with a greater amount of motions they represent. Average pair-wise divergence is included at the top of each bar.

3 DOF (wrist only) clusters are summarized as motions types, such as supination or deviation, referring to the most significant degree(s) of freedom. The dart-throwing motion (DTM), a hybrid of flexion and ulnar deviation, which has been described as a more stable and controllable axis of rotation [43], is re-discovered in our analysis as the average of the $2^{nd}$ cluster. This characteristic wrist motion has also been speculated as a key adaptation for tool-making in early hominids [44]. Since dendrogram interpretation is limited without animation, and while cluster descriptions for all three models are generalized in Fig. 5, readers are urged to view the average motions in the multi-media that accompanies this paper.

The chosen divergence measure and clustering algorithm outperformed Bézier and K-medoids methods at almost every number of clusters, reassuring its selection. The performance of K-medoids did not monotonically decrease with added clusters due to the algorithm reaching local minimums despite multiple iterations. Using Bézier coefficients to measure similarities between motions performed worse than DTW likely due to Bézier coefficients merely approximating the data whereas DTW takes the full joint angle trajectories into account and thus calculates a more representative divergence value.

Average pair-wise divergence and fPCA analysis capture the spread of a cluster and the directions of that spread, respectively. Although some clusters require as many as 8 fPC's to describe 80% of the variation, if the average pair-wise divergence is small, this does not necessarily mean that all of those fPC's are required to accurately reconstruct the motions, since they are largely similar to one another. The torso could potentially compensate for the variation as well.

The demonstrated cluster average in Fig. 8 and Fig. 9 can be directly implemented in a semi-autonomous robotic or prosthetic upper-limb model. The accompanying principal components in the same figures indicate how these motions vary, but can also be used to inform how to dynamically tune the trajectory to compensate for the motion variation within the cluster. This can be an indispensable aspect of control when, for example, reaching locations occur in continuous space. Future work should take advantage of fPCA findings in implementation of motion control and online adjustments.

If a common set of feature variables is identified, comparison may potentially be made with cyclical motions as well. One challenge, other than the small amplitudes of motion, is that cyclical motions do not have well defined start and end locations, and therefore rely on alternative representation methods such as wavelet transform or discrete Fourier transform [45]. However, these methods would not be appropriate for the type of data considered thus far in this study because reaching and transferring motions are seldom cyclical.

The decision to use joint angle data as the feature vector for this paper largely relied on the ability of recorded motions to be easily interpreted across individuals and its low dimensional representation. However, this choice suffers from giving each joint angle an equal weight when calculating the divergence between motions, while it may have been less of an issue for Cartesian coordinates of the upper-limb segments. Additionally, proximity to the discontinuities in two of the shoulder joint angles may cause them to have a larger impact when measuring motion similarity since the angle range is likely to be greater than for the other joint angles. Alternative arm features have been proposed in the literature, such as the arm triangle [46], or defining a new angle eliminating one of the discontinuities [47], either of which could be used in future iterations. Finally, although the decision to analyze the 3, 4, and 7 DOF arm models is relevant in a variety of applications, the methodology can be extended to alternative systems, such as to a full body kinematic chain.

## VI. CONCLUSION AND FUTURE WORK

This paper described a method that categorizes human arm motion during the performance of ADL tasks. Using data driven techniques to measure similarity between motions, average, and cluster, 11 motion categories were identified for the 7 DOF arm and 4 DOF elbow-wrist models and 5 motion categories for the 3 DOF wrist model. These clusters can be distinguished primarily based on start and end configurations of motions, further differentiated by specific types of manipulation.

The results align with intuition as well, making the proposed method a good candidate to describe other multi-DOF time-series systems. The application of this work is not task specific and is not exhaustive of the full set and complexity of motions within each task category, but instead provides a general framework that may be either applied in its current form for general use, improved on using fPCA, or could further be adapted to task specific scenarios to increase motion specificity.

An example includes obtaining a partial hierarchy of motions exclusively for feeding [25]. The proposed approach could also be applied to a subset of the presented data, such as decoupling the reaching location from the wrist orientation. Future developments include testing and verifying the identified average motions, implementation of a dynamic control of the average motions according to fPCA results, and identifying the role the torso plays during similar ADL tasks at different locations with respect to a fixed body frame.